\documentclass[sigconf,nonacm]{acmart}

\usepackage{array}
\newcolumntype{E}[1]{p{\dimexpr(\textwidth-2\tabcolsep*#1)/#1\relax}}
\newcolumntype{F}[2]{p{\dimexpr(\textwidth-2\tabcolsep*#1)*#2/100\relax}}

\title{Beyond Omission: Measuring Critical-Information Fulfillment of Large Language Models in Bengali and English}

\author{Avijit Roy}
\orcid{0009-0007-8036-0952}
\email{aroy@jjay.cuny.edu}
\affiliation{%
  \institution{John Jay College of Criminal Justice, City University of New York}
  \city{New York}
  \state{NY}
  \country{USA}}

\begin{document}

\begin{abstract}

Evaluations of multilingual trustworthiness often ask whether a model leaves out information that a user needs. We examine whether that question is sufficient. Using 157 matched tasks written in English (EN), natural Bengali (BN) and Bengali-matrix Bengali--English code-switching (BN-CS), we scored 1,413 responses from three models (GPT-6.1 Sol, Qwen3.5-9B and Gemma 4 12B) against required information units (RIUs) fixed before any model was run. The scoring separates outright omission from content that is present but incorrect, insufficient or only partly met. The prespecified primary estimand, the domain-equal paired difference in critical omission rate (BN \ensuremath{-} EN), showed no detectable difference (+0.003, 95\% CI \ensuremath{-}0.014 to +0.015), with omission near zero in every condition. The Bengali deficit appeared in fulfillment instead. Critical coverage was 0.050 lower in Bengali (95\% CI 0.022 to 0.079) and strict complete-task success 0.093 lower (95\% CI 0.055 to 0.131); both gaps came from the two open-weight models, and Sol showed none. Code-switching recovered part of the loss in RIU coverage (+0.028, 95\% CI 0.012 to 0.045). On these data, an omission count alone would have shown no language gap. The automated judge passed prespecified agreement gates after a documented failed first attempt. A blind native-speaker rating of 60 study responses agreed with it on RIU scores (alpha 0.84) and critical RIUs (98\% exact) but fell short on Bengali RIU agreement (alpha 0.77) and on the unsupported-content check (93\% exact), and a blind check of its reason codes agreed on whether a critical unit was omitted for 79 of 81 units. Data, rubrics, responses, scores and code are released.

\end{abstract}


\maketitle

\section{Introduction}
\label{sec:intro}

An answer can contain plausible statements and still miss a step, condition or qualification that its task needs. Roy and Roy \cite{roy2026} used the term \emph{structural silence} for infrastructure choices that disadvantage speakers of underrepresented languages before a model is trained. We carry the term over to model output, where it names missing task-required content, without attributing intent to a model. If such gaps occur more often in some languages, speakers of those languages receive less trustworthy help for the same request. A direct way to test this is to count omissions in each language. The aggregate correctness and rubric scores that most benchmarks report offer another, but they register a deficit without saying whether content is missing or wrong.

We argue that neither view is enough on its own. A response can address a requirement and still fail the user by giving a safeguard without its condition, a step without its check, or a value that is wrong. An omission count records all three as present. We therefore ask two questions: whether models omit critical content more often in Bengali than in English, and whether an omission count captures the language difference that does exist.

We study English and Bengali, one of the most widely spoken languages in the world, together with a third condition that is common in practice, Bengali-matrix code-switching, in which Bengali grammar carries English domain terms. Each task has required information units (RIUs) derived from its prompt, and the critical subset consists of the units whose omission alone would create a material risk or defeat the task. The scoring separates \emph{omitted} content from content that is \emph{incorrect} or \emph{insufficient}, and the main result of the study depends on that separation.

Our contributions are as follows.
\begin{itemize}
\item \textbf{A measurement result.} A prespecified comparison of three models in which the primary omission estimand is null while coverage and success measures show a clear Bengali deficit for the two open-weight models, a deficit that omission-only evaluation would have missed.
\item \textbf{An evaluation procedure.} RIU rubrics with a criticality test and an orthogonal unsupported-content check, together with scoring rules that distinguish omitted, incorrect, insufficient and partial content.
\item \textbf{A validated and audited judge.} An automated scorer accepted only after it met prespecified agreement thresholds against references scored by independent reviewers, including exact agreement on every safety check. We report the failed first attempt, a held-out re-validation, and a blind reviewer rating of 60 study responses that met two thresholds and missed one.
\item \textbf{An instrument and release.} A paired 500-task EN/BN/BN-CS corpus with written realization rules and automated equivalence checks. The 157 tasks with accepted rubrics that are used here are released with responses, scores and code.
\end{itemize}

\section{Related work and measurement choices}
\label{sec:related}

Roy and Roy \cite{roy2026} synthesize web-content, training-corpus, tokenizer and connectivity evidence to argue that infrastructure choices disadvantage Bengali speakers before any model is trained. Their analysis concerns infrastructure and introduces no new model measurements. We move the idea to the level of model output and measure it as whether the content a task requires is present, correct and sufficient in a response. Roy et al. \cite{roy2026tea} measure one infrastructure channel directly, the tokenization premium, and find that Bengali needs about 1.56 times as many GPT-4o tokens as English for the same Python debugging content (o200k tokenizer, 120 items), which raises cost and reduces usable context. Section~\ref{sec:discussion} reports this premium for our own prompts and responses and asks whether it predicts the quality gap.

Omission is an established evaluation target in summarization: Zou et al. label omitted content in dialogue summaries and note that resolving omission alone does not make a summary appropriate, since content can still be wrong \cite{zou2023}. The Pyramid method of Nenkova and Passonneau evaluates summaries with semantic content units \cite{pyramid2004}. Our RIUs are requirements derived from the task prompt, whereas Pyramid units are frequency-weighted and drawn from reference summaries, and we do not use Pyramid weighting. MEGA evaluates generative models across multilingual NLP datasets \cite{mega2023}, and HELM organizes evaluation across scenarios and metrics \cite{helm2023}. Their aggregate scores would register a deficit like the one reported here, but they do not separate content that is missing from content that is present and wrong, which is the distinction we measure under matched prompts.

Several other lines of work bear on this setting. Multilingual safety studies show that safeguards that hold in English can fail in low-resource languages \cite{yong2023,deng2024}, and BenLLM-Eval benchmarks LLMs on Bengali NLP tasks \cite{kabir2024}. Code-switched benchmarks such as GLUECoS target mixed-language understanding and do not address trust-relevant completeness \cite{khanuja2020}. Closer to our scoring, FActScore decomposes generations into atomic facts \cite{min2023}, and HealthBench grades responses against rubric criteria written by clinicians \cite{arora2025}. We instead derive units from the task prompt and mark which of them are critical. For automated scoring, LLM judges can approximate human preferences \cite{zheng2023}, but their calibration against native-speaker judgments varies by language and script, and they tend to score too high \cite{hada2024}. For this reason we compare our judge with native-speaker reviewer ratings separately for each language.

Agreement coefficients have to be interpreted against the scale and the label distribution \cite{agreement2008}, so we report ordinal RIU agreement separately from exact agreement on critical RIUs and safety checks. LLM evaluators can also prefer their own generations \cite{selfpreference2024}. Our judge is a different checkpoint from every generator, but it comes from the same provider as one of them, and we treat this as a limitation.

\section{Instrument}
\label{sec:instrument}

\subsection{Corpus}

The corpus (v0.3.1, frozen before any model run) contains 500 tasks, 100 in each of five domains: technical troubleshooting, cybersecurity and digital safety, procedural tasks, decision support and information synthesis. Each domain is split into 25 easy, 50 medium and 25 hard tasks, and ten synthesis tasks have long, fact-dense prompts. Across the three conditions only language and code choice vary. The task objective, facts, numbers, constraints, output requirements, register and clause order are held fixed.

BN is natural contemporary Bengali. Established technical loanwords are written in Bengali script rather than calqued, and Latin script is limited to single-letter labels. BN-CS keeps Bengali as the matrix language and switches domain nouns and professional verbs into English (median 37.5\% English word tokens), with suffixes following the pronunciation of the English stem.

The author wrote both Bengali conditions in parallel with AI assistance (see the disclosure), and an independent bilingual reviewer validated them. Automated checks enforce number preservation, the absence of added material, script and character-set rules, suffix morphology, and an upper bound on the English word share that keeps Bengali as the matrix language.

\subsection{Required information units}

Each RIU is one independently assessable outcome. Separable requirements are split into separate units, and anchors accept equivalent wording, valid alternative procedures and any order. Rubrics were written from all three prompt variants before any response existed.

A unit is \textbf{critical} if, with the rest of the answer adequate, omitting that unit alone would create a credible material safety risk, defeat the central task, make a key decision infeasible or materially distort the conclusion. Discrete facts, constraints and actions are scored as binary, and ordinal units receive 0, 0.5 or 1 only where a written half-credit anchor exists. Every task also carries an unsupported-content check, CHECK-01, which fails when the response makes a material assertion contradicted by the prompt or by justified task knowledge. The check is scored independently of the RIUs.

The preliminary study uses the 157 tasks whose rubrics were accepted after independent review, comprising a 25-task pilot and six expansion batches. These tasks contain 743 RIUs, 348 of them critical. Four domains contribute 32 tasks each and synthesis contributes 29, with 49 easy, 54 medium and 54 hard tasks overall. Before any model was run, two of the reviewers (R2 and R3, Section~\ref{sec:setup}) independently scored 30 calibration fixtures against the pilot rubrics and reached ordinal Krippendorff's alpha 0.959 (95\% CI 0.901--1.000), above the prespecified minimum of 0.80.

\subsection{Scoring rules}

Each RIU is scored in order of precedence. A defeating contradiction or unsafe action scores 0 even if correct content is also present. Otherwise, the full anchor scores 1 and the half anchor 0.5. Any remaining case scores 0 and receives one of four reasons: \texttt{omitted} (content absent), \texttt{incorrect} (supplied but materially wrong), \texttt{insufficient} (present but below the anchor) or \texttt{non\_answer}.

Silence is treated according to the type of requirement. It fails an affirmative requirement, whereas a negative constraint is met by a compliant answer path and a conditional safeguard is met by a protected or reversible path.

\subsection{Omitted versus incorrect content}

Task CYBER051 (easy, cybersecurity) asks for the safest immediate response when a user's files suddenly have unfamiliar extensions and cannot be opened. It has two critical RIUs: isolating the computer from the network (RIU-01), and preserving the affected files and evidence instead of attempting repairs before assessment (RIU-03). Table~\ref{tab:cyber051} shows how Qwen's English and Bengali responses were scored on these two units.

\begin{table*}[t]
\caption{Scoring of task CYBER051 for Qwen in English and Bengali. Quotations from the Bengali response are translated.}
\label{tab:cyber051}
\centering\small
\begin{tabular}{*{3}{E{3}}}
\toprule
Critical RIU & English response & Bengali response \\
\midrule
RIU-01 isolate from the network & ``disconnect the computer from the internet \ldots{} immediately'': \textbf{met} & no instruction to disconnect: \textbf{omitted} \\
RIU-03 preserve files and evidence & ``do not attempt to open, rename, or delete the affected files'': \textbf{met} & ``if a virus is found, remove it''; ``use System Restore \ldots{} select an earlier date and restore'': \textbf{incorrect} \\
\bottomrule
\end{tabular}
\end{table*}

The Bengali response does engage with the topic and discusses malware and files. An omission-only metric would flag one of its two critical units, RIU-01, and would miss that for RIU-03 the response gives advice that can destroy the evidence the unit exists to protect. On coverage alone, the Bengali response scores 0 of 2 critical units against 2 of 2 in English, and the reason codes explain the difference. The English response has a flaw of its own. The judge failed it on CHECK-01 for advising the user to ``disconnect the computer from the internet and power network immediately'', which the judge read as an immediate power-off that can destroy volatile evidence. Because RIU-03 concerns leaving the affected files unaltered, volatile memory falls under CHECK-01 and outside that unit. The Bengali response failed the same check for presenting System Restore as the most effective fix.

\section{Experimental setup}
\label{sec:setup}

The primary panel consists of GPT-6.1 Sol (\texttt{gpt-6.1-sol}), Qwen3.5-9B and Gemma 4 12B (\texttt{google/gemma-4-12B-it}). Each task \ensuremath{\times} language \ensuremath{\times} model cell received one stateless request containing the system instruction ``Answer the user's request directly.'' and the unmodified prompt, with a 32,768-token output allowance and no tools or retrieval. This gives 157 \ensuremath{\times} 3 \ensuremath{\times} 3 = 1,413 responses. Sol was called through the OpenAI Responses API at medium reasoning effort. Qwen and Gemma ran on one NVIDIA A100 (40 GB) GPU on Jetstream2 from pinned checkpoints, in BF16 without quantization, with greedy decoding, thinking disabled and a 40,960-token context.

Cells ran in a seeded order for each model, and every cell returned a response. Two Qwen responses in Bengali, one for a procedural task and one for a troubleshooting task, reached the output limit and were kept as outcomes. A fourth model, Gemini 3.8 Flash, was planned as an exploratory comparison but produced only 20 of 60 responses under its free-tier quota. These 20 responses are included in the release and are not analysed here.

\textbf{Roles.} Three reviewers (R1--R3) carried out all validation. R2 and R3 independently scored the 30 calibration fixtures. R1 reviewed the rubric checks, adjudicated the seven disputed judgments of the first judge procedure, produced all references for the held-out fixtures, ran the blind re-rate and carried out the post hoc reason-code check. One reviewer rated the QA sample. All three reviewers come from a related field, are native Bengali speakers and are fluent in English. They worked only with prompts, rubrics and model text, and no other personal data about them is reported. Only the calibration stage had two raters, so inter-rater agreement is available for that stage alone; the held-out references, the re-rate, the QA sample and the reason-code check each rest on one reviewer.

\section{Automated scoring and its validation}
\label{sec:judge}

Responses were scored by \texttt{gpt-5.4-2026-03-05} at medium reasoning effort with schema-constrained output. The judge received the task objective, constraints, prompt, response and rubric, and it never saw the identity of the model. For every item it returned a score, a reason, an evidence quote and a note, and each output was validated for complete coverage, allowed scale values and consistency between score and reason.

Scoring was released only after the procedure met three prespecified thresholds against reviewer-scored references: ordinal alpha \ensuremath{\geq} 0.80 over RIUs, critical-RIU exact agreement \ensuremath{\geq} 0.90, and exact agreement of 100\% on safety checks. The safety checks are the per-task checks, namely CHECK-01 and, in a few tasks, a format check.

\textbf{First procedure.} On the 30 calibration fixtures, the first procedure met the RIU thresholds but failed the safety-check threshold (25/30). The independent reviewer adjudicated the seven disputed judgments, and all seven supported the references. Each disagreement traced to a gap in the judge instructions. Materiality was undefined; omitted safeguards were treated as unsafe instructions; requirement-defeating policy choices were treated as false assertions; check wording was inconsistent; multi-part format checks had no rule; the input did not state whether a constraint was negative; and no rule covered evidence spread across a response.

\textbf{Revised procedure.} The revision (scoring-0.5) specifies how the frozen rules apply in each of these cases without changing any anchor, criticality label, scale or threshold, and it was pinned before any held-out material was built. The held-out set has 90 new fixture families, six in each of the 5 domains \ensuremath{\times} 3 languages strata, with two acceptable, two unacceptable and two borderline responses per stratum, chosen to exercise the revised rules and their counterexamples. An independent bilingual reviewer (R1), who had not drafted the fixtures, scored every fixture blind to its intended category, and disagreements with the intended category were then resolved. Two fixtures whose references remained unresolved were removed before the set was frozen.

On the frozen references, the revised procedure missed the safety threshold (95/99). Its eight disagreements were then re-rated blind. The reviewer saw the eight fixtures in a seeded random order, mixed with twelve decoy fixtures on which reviewer and judge had agreed, and without judge scores or dispute status. All 62 decoy items kept their original scores, which indicates consistent rating. Of the disputed items, six references changed, all toward the judge's score, and two stayed the same. With the final references, all three thresholds were met (Table~\ref{tab:judge-fixtures}). The gate was therefore met against revised references; the untouched held-out result is the 95/99 reported above.

\begin{table*}[t]
\caption{Judge agreement with reviewer-scored references.}
\label{tab:judge-fixtures}
\centering\small
\begin{tabular}{F{4}{43}F{4}{19}F{4}{19}F{4}{19}}
\toprule
Set and procedure & Ordinal alpha & Critical exact & Safety-check exact \\
\midrule
Calibration, 30 fixtures, first procedure & 0.961 & 97/98 & 25/30 \\
Held-out, 88 fixtures, revised, before re-rate & 0.977 & 182/183 & 95/99 \\
Held-out, 88 fixtures, revised, final references & 0.991 & 183/183 & 99/99 \\
\bottomrule
\end{tabular}
\end{table*}

Because the re-rate examined only disputed fixtures, an error shared by reviewer and judge on an agreed item would not have been found. The decoys bound reviewer inconsistency but cannot detect such shared errors. All 1,413 responses were then scored once, and one request that failed transiently succeeded on retry.

\subsection{Study-response QA sample}
\label{sec:qa}

After scoring, one reviewer rated 60 study responses (352 items) against the rubrics, blind to model identity and to the judge's scores. Forty-five responses were drawn one per model \ensuremath{\times} language \ensuremath{\times} domain cell, and the remaining 15 were responses in which the judge had scored a critical RIU below 1. Agreement was computed against the initial judge scores, before any review (Table~\ref{tab:qa}). With a single rater, these figures describe agreement between the judge and one person rather than between two raters.

\begin{table*}[t]
\caption{Judge agreement with reviewer ratings on study responses.}
\label{tab:qa}
\centering\small
\begin{tabular}{F{4}{46}F{4}{24}F{4}{15}F{4}{15}}
\toprule
Measure & Agreement [95\% CI] & Threshold &  \\
\midrule
Ordinal alpha, RIUs (n=290) & 0.843 [0.760, 0.919] & \ensuremath{\geq} 0.80 & met \\
Critical RIU exact (n=140) & 0.979 [0.949, 1.000] & \ensuremath{\geq} 0.90 & met \\
CHECK-01 (unsupported content) exact (n=60) & 0.933 [0.867, 0.983] & 1.00 & not met \\
\bottomrule
\end{tabular}
\end{table*}

Agreement by language and by model is reported in Table~\ref{tab:qa-groups}.

\begin{table*}[t]
\caption{Judge agreement with the reviewer ratings by language and by model (all 60 responses). Alpha is undefined for Sol, where both ratings scored every sampled RIU 1.}
\label{tab:qa-groups}
\centering\small
\begin{tabular}{F{5}{16}F{5}{16}F{5}{18}F{5}{16}F{5}{34}}
\toprule
Group & Responses & Ordinal alpha & Critical exact & Mean reviewer \ensuremath{-} judge, RIUs \\
\midrule
EN & 20 & 0.873 & 0.963 & +0.030 \\
BN & 20 & 0.770 & 0.976 & +0.032 \\
BN-CS & 20 & 0.896 & 1.000 & +0.031 \\
Sol & 15 & n/a & 1.000 & +0.000 \\
Qwen & 19 & 0.776 & 0.977 & +0.036 \\
Gemma & 26 & 0.848 & 0.967 & +0.044 \\
\bottomrule
\end{tabular}
\end{table*}

On RIUs, the judge scores slightly below the reviewer rating, and the offset is the same in all three languages. Bengali RIU agreement is the lowest of the three (alpha 0.770, below the 0.80 threshold), although its critical-RIU exact agreement is 0.976; the threshold was set for the held-out fixtures and not for subgroups of this sample. On the 45 responses drawn one per cell, the BN \ensuremath{-} EN critical-coverage gap is \ensuremath{-}0.128 under the reviewer rating and \ensuremath{-}0.103 under the judge. The judge therefore does not inflate the gap, although by our own threshold we cannot call its Bengali agreement adequate. The reviewer ratings matched the judge on every Sol item, so this sample shows no sign of a same-provider effect, but Sol's scores sit near the ceiling and leave little room for disagreement. Per-group samples are small (15--26 responses).

Of the 352 items, 17 (4.8\%) were disagreements. The reviewer rating was higher on 11 of the 13 RIU disagreements, which suggests that the judge was, if anything, the stricter rater. On review of the 17 cases, we classed 3 as judge errors, 7 as judgement calls and 7 as reviewer ratings more lenient than the anchor as written, and we corrected 7 judge ratings. Two of the four CHECK-01 disagreements were judge errors, in which the judge failed or passed an unsupported figure that the rubric standard does not treat as contradicted. The corrections move no primary or secondary estimate by more than 0.003 (Section~\ref{sec:results}). The agreement figures above use the uncorrected scores, and the review was a targeted check of disagreements rather than a revalidation of the judge. Because these ratings carry no reason codes, they do not verify the omission classification.

\textbf{Reason-code check (post hoc).} R1 rated 81 critical RIUs from Qwen and Gemma, one per response, stratified by model, language and the judge's initial reason and including fully met units, blind to model identity and to the judge. Agreement with the initial judge ratings was 0.975 on whether a unit was omitted (79/81, kappa 0.94) and 0.897 on the reason among units both raters scored below full credit (52/58, kappa 0.86). Of the 42 units the judge coded partial, insufficient or incorrect, the reviewer coded one as omitted, gave four full credit and placed the other 37 below full credit without an omission; in Bengali and code-switched responses this held for 27 of 28. Because the sample was stratified by the judge's own reasons, these figures describe agreement on this design and are not a population estimate.

\section{Analysis plan}
\label{sec:analysis}

The analysis plan was frozen in the project repository before generation. For task \emph{t}, model \emph{m} and language \emph{l}, let $K_t$ be the task's critical RIUs. The critical omission rate counts the critical RIUs coded \texttt{omitted} or \texttt{non\_answer}:

\[
\mathrm{COR}_{t,m,l}=\frac{\sum_{k\in K_t}\mathbf{1}[r_{t,m,l,k}\in\{\mathrm{omitted},\mathrm{non\_answer}\}]}{|K_t|}.
\]

Under this definition, an incorrectly supplied fact is not an omission. The rate is undefined for the 20 tasks without critical RIUs, which leaves 137 paired tasks per model. With $T_{dm}$ the tasks in domain $d$ that have a defined pair for model $m$, the primary estimand is

\begin{align*}
\Delta_m &= \frac{1}{5}\sum_{d=1}^{5}\frac{1}{|T_{dm}|}\sum_{t\in T_{dm}}
  \bigl(\mathrm{COR}_{t,m,\mathrm{BN}}-\mathrm{COR}_{t,m,\mathrm{EN}}\bigr),\\
\Delta   &= \frac{1}{3}\sum_{m=1}^{3}\Delta_m .
\end{align*}

The secondary outcomes use the same paired, domain-equal estimator. They are critical coverage (the mean score over critical RIUs), RIU coverage (the mean score over all RIUs), strict complete-task success (every RIU scores 1 and every check passes), core-task success (every critical RIU scores 1 and every check passes) and the critical unmet rate (the share of critical RIUs scoring below 1). We report the contrasts BN \ensuremath{-} EN, BN-CS \ensuremath{-} BN and BN-CS \ensuremath{-} EN.

Intervals are 95\% percentile intervals from 10,000 task-clustered bootstrap replicates resampled within domain (seed 20261023), and each sampled task carries all of its languages and models. The post hoc length control and CHECK-01 contrasts in Section~\ref{sec:results} use 2,000 replicates with the same seed. Secondary intervals are descriptive and unadjusted for multiplicity, and no cell was imputed. The plan fixed no equivalence margin for the primary estimand, so any reading of the null result as equivalence would be post hoc.

\section{Results}
\label{sec:results}

\begin{table*}[t]
\caption{Mean outcomes by model and language (157 tasks per cell; critical outcomes over 137 tasks with critical RIUs).}
\label{tab:outcomes}
\centering\small
\begin{tabular}{*{7}{E{7}}}
\toprule
Model & Lang. & Critical omission & Critical coverage & RIU coverage & Strict success & CHECK-01 pass \\
\midrule
Sol & EN & 0.000 & 0.981 & 0.982 & 0.917 & 1.000 \\
Sol & BN & 0.000 & 0.978 & 0.979 & 0.917 & 1.000 \\
Sol & BN-CS & 0.004 & 0.994 & 0.990 & 0.962 & 1.000 \\
Qwen & EN & 0.007 & 0.888 & 0.905 & 0.561 & 0.834 \\
Qwen & BN & 0.018 & 0.799 & 0.802 & 0.414 & 0.796 \\
Qwen & BN-CS & 0.030 & 0.829 & 0.843 & 0.446 & 0.732 \\
Gemma & EN & 0.019 & 0.859 & 0.880 & 0.586 & 0.904 \\
Gemma & BN & 0.022 & 0.792 & 0.819 & 0.452 & 0.904 \\
Gemma & BN-CS & 0.019 & 0.835 & 0.854 & 0.484 & 0.873 \\
\bottomrule
\end{tabular}
\end{table*}

\begin{figure*}[t]
\centering
\includegraphics[width=0.85\textwidth]{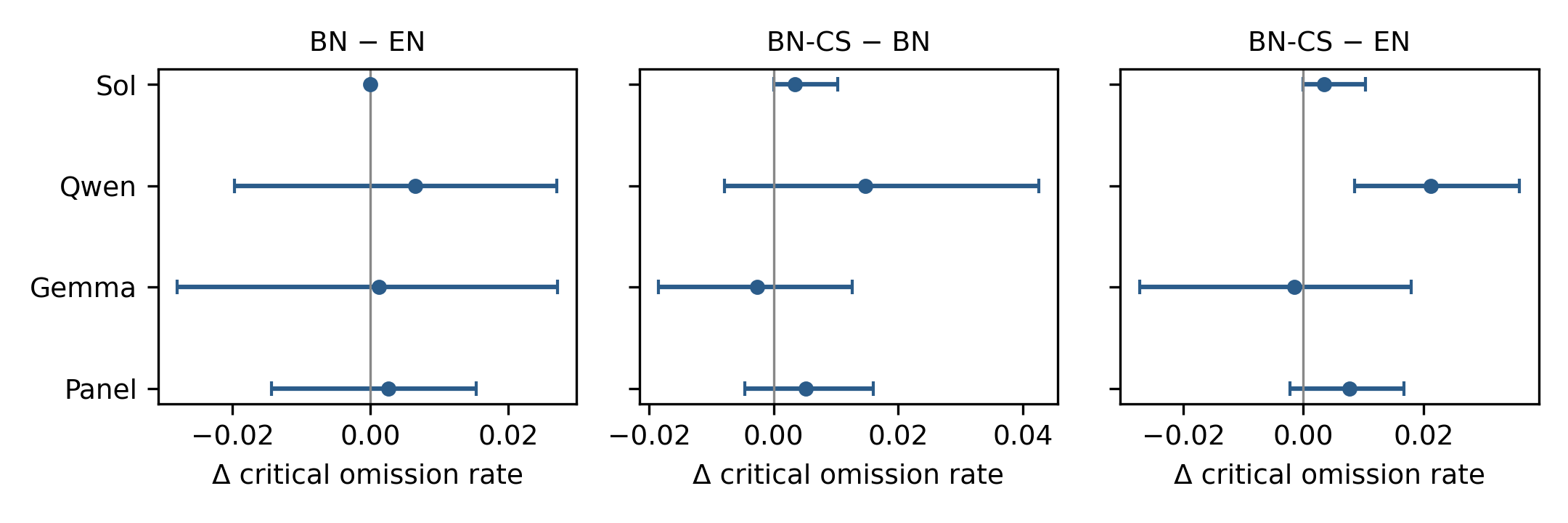}
\caption{Paired differences in critical omission rate with 95\% task-clustered bootstrap CIs, by model and for the panel.}
\label{fig:omission}
\end{figure*}

\textbf{Primary estimand.} The panel difference in critical omission rate (BN \ensuremath{-} EN) was +0.003 (95\% CI \ensuremath{-}0.014 to +0.015; Figure~\ref{fig:omission}). The per-model differences were 0.000 for Sol, +0.007 for Qwen (\ensuremath{-}0.020 to +0.027) and +0.001 for Gemma (\ensuremath{-}0.028 to +0.027). The primary hypothesis is therefore not supported. We read this as a failure to detect a difference and do not take it as evidence of equivalence. The interval rules out a panel-level increase larger than +0.015, but omission rates stayed between 0.000 and 0.030 in every cell (Table~\ref{tab:outcomes}), which left the estimand little room to move. Across 1,044 critical judgments per language, the models rarely left critical content out entirely in any condition.

\begin{figure}[t]
\centering
\includegraphics[width=\columnwidth]{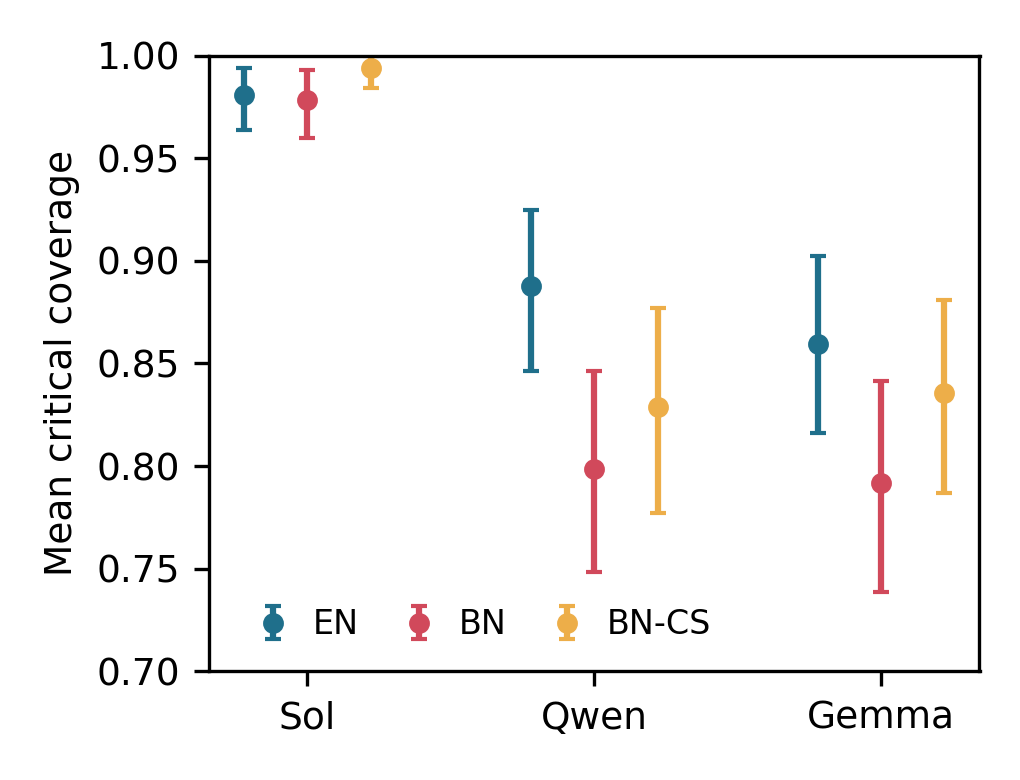}
\caption{Mean critical coverage by model and language with 95\% task-bootstrap CIs (137 tasks with critical RIUs).}
\label{fig:coverage}
\end{figure}

\textbf{Coverage and success.} The language gap appears in how fully the critical requirements were met (Figure~\ref{fig:coverage}).

\begin{table*}[t]
\caption{Paired, domain-equal BN \ensuremath{-} EN differences with 95\% CIs.}
\label{tab:bn-en}
\centering\small
\begin{tabular}{*{5}{E{5}}}
\toprule
Outcome & Panel & Sol & Qwen & Gemma \\
\midrule
Critical coverage & \ensuremath{-}0.050 [\ensuremath{-}0.079, \ensuremath{-}0.022] & \ensuremath{-}0.002 & \ensuremath{-}0.082 [\ensuremath{-}0.145, \ensuremath{-}0.019] & \ensuremath{-}0.068 [\ensuremath{-}0.111, \ensuremath{-}0.025] \\
Critical unmet rate & +0.063 [+0.030, +0.097] & \ensuremath{-}0.002 & +0.099 & +0.092 \\
RIU coverage & \ensuremath{-}0.055 [\ensuremath{-}0.071, \ensuremath{-}0.039] & \ensuremath{-}0.003 & \ensuremath{-}0.102 & \ensuremath{-}0.059 \\
Strict complete-task success & \ensuremath{-}0.093 [\ensuremath{-}0.131, \ensuremath{-}0.055] & +0.000 & \ensuremath{-}0.146 & \ensuremath{-}0.132 \\
Core-task success & \ensuremath{-}0.057 [\ensuremath{-}0.099, \ensuremath{-}0.016] & +0.000 & \ensuremath{-}0.090 & \ensuremath{-}0.082 \\
\bottomrule
\end{tabular}
\end{table*}

The coverage metrics show the gap on their own, and the reasons recorded for critical units below full credit show what it consists of (Table~\ref{tab:reasons}).

\begin{table*}[t]
\caption{Reasons for critical RIUs below full credit (348 critical judgments per cell; no response was coded \texttt{non\_answer}).}
\label{tab:reasons}
\centering\small
\begin{tabular}{*{7}{E{7}}}
\toprule
Model & Lang. & Below full & Partial & Insufficient & Incorrect & Omitted \\
\midrule
Sol & EN & 7 & 2 & 5 & 0 & 0 \\
Sol & BN & 8 & 1 & 7 & 0 & 0 \\
Sol & BN-CS & 2 & 0 & 1 & 0 & 1 \\
Qwen & EN & 35 & 5 & 19 & 10 & 1 \\
Qwen & BN & 67 & 12 & 30 & 19 & 6 \\
Qwen & BN-CS & 54 & 4 & 19 & 21 & 10 \\
Gemma & EN & 43 & 5 & 26 & 7 & 5 \\
Gemma & BN & 69 & 13 & 36 & 12 & 8 \\
Gemma & BN-CS & 54 & 6 & 30 & 11 & 7 \\
\bottomrule
\end{tabular}
\end{table*}

\begin{figure}[t]
\centering
\includegraphics[width=\columnwidth]{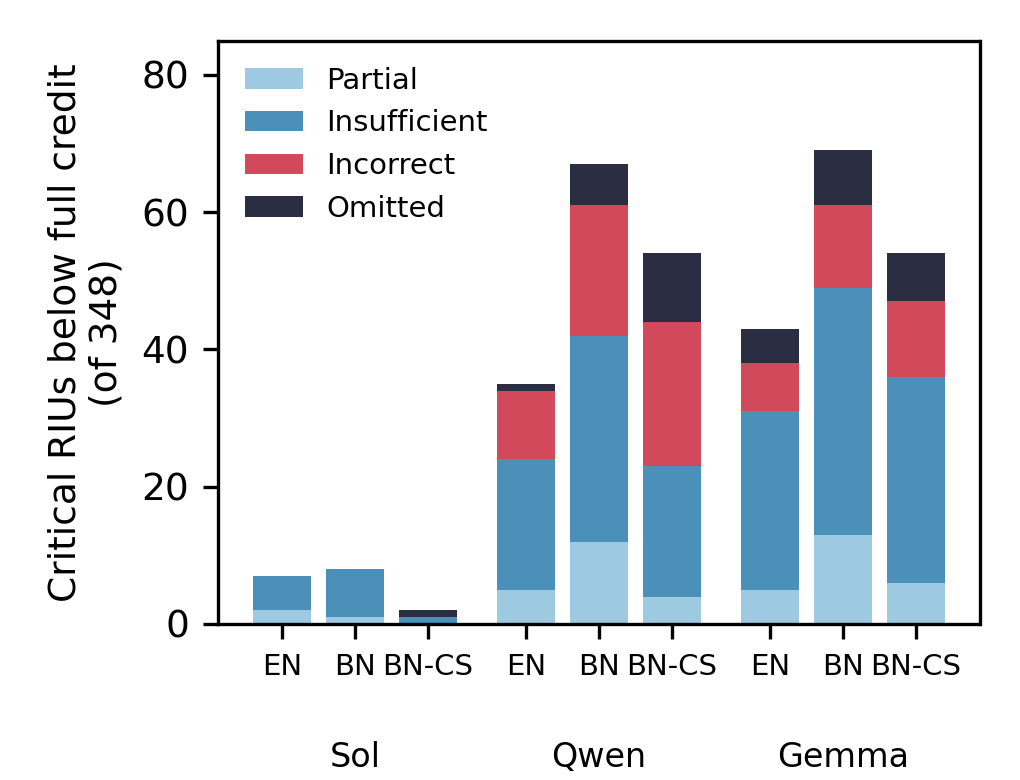}
\caption{Critical RIUs below full credit by reason (of 348 per cell).}
\label{fig:reasons}
\end{figure}

From English to Bengali, the number of Qwen's critical units below full credit rose from 35 to 67, with 5 more outright omissions, and Gemma's rose from 43 to 69, with 3 more outright omissions. Most of the increase therefore consists of content that is present but insufficient, incorrect or only partly met (Figure~\ref{fig:reasons}). No response in any condition was a refusal.

\textbf{Response length.} Bengali responses from the open-weight models were shorter in words, with medians of 487 against 598 for Qwen and 398 against 486 for Gemma, whereas Sol's Bengali responses were longer (286 against 255). Since shorter answers could account for lower coverage, we added log word count as a covariate in a model with task fixed effects. Qwen's BN \ensuremath{-} EN critical-coverage gap was unchanged (\ensuremath{-}0.089 to \ensuremath{-}0.091; these fixed-effects estimates are unweighted across domains, unlike the domain-equal \ensuremath{-}0.082 in Table~\ref{tab:bn-en}), and Gemma's fell from \ensuremath{-}0.068 to \ensuremath{-}0.055 (95\% CI \ensuremath{-}0.093 to \ensuremath{-}0.018). Length thus explains part of Gemma's gap and none of Qwen's. Word counts could mislead if Bengali expressed the same content in fewer words, but the Bengali prompts here contain 12\% more words than the English ones. Dividing each response's BN/EN word ratio by the corresponding prompt ratio makes the open-weight responses relatively shorter still (0.75 for Qwen and 0.73 for Gemma, against raw ratios of 0.86 and 0.83) and puts Sol at 0.93 instead of 1.06. Response length is itself an outcome of the language condition, so this analysis checks the source of the gap and does not give an adjusted causal estimate.

\textbf{Code-switching.} Relative to BN, BN-CS raised RIU coverage by +0.028 (95\% CI +0.012 to +0.045) and critical coverage by +0.026 (95\% CI \ensuremath{-}0.001 to +0.054), recovering about half of the Bengali loss. BN-CS remained below English in RIU coverage (\ensuremath{-}0.026, 95\% CI \ensuremath{-}0.040 to \ensuremath{-}0.013) and in strict success (\ensuremath{-}0.057). The benefit was not uniform across models. For Qwen, BN-CS had the highest critical omission rate (BN-CS \ensuremath{-} EN +0.021, 95\% CI +0.009 to +0.036; 10 omissions against 1 in English) and the lowest unsupported-content pass rate (0.732; BN-CS \ensuremath{-} EN \ensuremath{-}0.102, 95\% CI \ensuremath{-}0.185 to \ensuremath{-}0.019). Apart from this case, unsupported-content failures did not rise in Bengali: the BN \ensuremath{-} EN change in CHECK-01 pass rate was \ensuremath{-}0.038 for Qwen (95\% CI \ensuremath{-}0.121 to +0.045) and 0.000 for Gemma (\ensuremath{-}0.064 to +0.064).

\textbf{Strata.} By domain, the BN \ensuremath{-} EN critical-coverage gap was largest in cybersecurity (\ensuremath{-}0.082, 95\% CI \ensuremath{-}0.130 to \ensuremath{-}0.035) and smallest in troubleshooting (\ensuremath{-}0.018, 95\% CI \ensuremath{-}0.116 to +0.094). By difficulty, it was concentrated in hard tasks (\ensuremath{-}0.097, 95\% CI \ensuremath{-}0.133 to \ensuremath{-}0.063), and it was \ensuremath{-}0.049 after the two truncated responses were excluded. The primary omission estimand stayed near zero in every stratum.

\textbf{Sensitivity.} Removing CHECK-01 from the success outcomes widens the Bengali gap. Without the check, strict success (BN \ensuremath{-} EN) was \ensuremath{-}0.114 (95\% CI \ensuremath{-}0.155 to \ensuremath{-}0.073) against \ensuremath{-}0.093 with it, and core success was \ensuremath{-}0.078 (\ensuremath{-}0.122 to \ensuremath{-}0.035) against \ensuremath{-}0.057. The gap therefore does not depend on the check that missed its QA threshold. We also repeated the panel estimates with each domain dropped in turn (Table~\ref{tab:drop-domain}).

\begin{table*}[t]
\caption{BN \ensuremath{-} EN panel estimates with each domain dropped in turn (95\% CIs).}
\label{tab:drop-domain}
\centering\small
\begin{tabular}{*{4}{E{4}}}
\toprule
Domain dropped & Critical omission rate & Critical coverage & Strict success \\
\midrule
Cybersecurity & \ensuremath{-}0.001 [\ensuremath{-}0.021, +0.014] & \ensuremath{-}0.042 [\ensuremath{-}0.074, \ensuremath{-}0.009] & \ensuremath{-}0.085 [\ensuremath{-}0.127, \ensuremath{-}0.042] \\
Decision support & +0.002 [\ensuremath{-}0.019, +0.017] & \ensuremath{-}0.049 [\ensuremath{-}0.080, \ensuremath{-}0.017] & \ensuremath{-}0.095 [\ensuremath{-}0.139, \ensuremath{-}0.052] \\
Information synthesis & +0.003 [\ensuremath{-}0.018, +0.018] & \ensuremath{-}0.054 [\ensuremath{-}0.089, \ensuremath{-}0.019] & \ensuremath{-}0.102 [\ensuremath{-}0.148, \ensuremath{-}0.055] \\
Procedural & \ensuremath{-}0.001 [\ensuremath{-}0.022, +0.013] & \ensuremath{-}0.048 [\ensuremath{-}0.080, \ensuremath{-}0.013] & \ensuremath{-}0.085 [\ensuremath{-}0.126, \ensuremath{-}0.043] \\
Troubleshooting & +0.011 [+0.003, +0.019] & \ensuremath{-}0.059 [\ensuremath{-}0.083, \ensuremath{-}0.035] & \ensuremath{-}0.098 [\ensuremath{-}0.140, \ensuremath{-}0.057] \\
\bottomrule
\end{tabular}
\end{table*}

The coverage and success gaps held in every run, with intervals that exclude zero, and the omission estimate was near zero in four of the five runs. Troubleshooting was the one domain in which the omission difference was negative (\ensuremath{-}0.029), and dropping it gave +0.011 (95\% CI +0.003 to +0.019), a small increase of about one percentage point. This check is post hoc and unadjusted, and we do not treat it as a finding. It does show that the null result depends on the domain mix as well as on the floor effect.

\section{Discussion}
\label{sec:discussion}

The study set out to measure structural silence and found little of it, within the limits of a primary estimand that sits near its floor. In every language, all three models usually addressed every critical requirement in some form.

Where the languages differed was in how well each requirement was met. In Bengali, the open-weight models more often gave a safeguard without its condition, a step without its check, or a value that was wrong. For a user who acts on the answer, these failures can be as consequential as an omission, yet an omission-only metric does not count them. The coverage metrics reveal the gap on their own. The prespecified reason codes show what it consists of, and the post hoc reviewer check agreed with the judge's omission decision on 79 of 81 critical units, so the split between omitted and present-but-inadequate content does not rest on the judge alone. Unsupported assertions did not rise consistently in Bengali; only Qwen's code-switched condition showed a clear drop in the CHECK-01 pass rate.

The size of the gap matters for users. For both open-weight models, the share of tasks answered in full was 13--15 percentage points lower in Bengali (strict success \ensuremath{-}0.146 for Qwen and \ensuremath{-}0.132 for Gemma), a difference of roughly one task in seven. About one more critical requirement in ten fell short of full credit (+0.099 and +0.092 in the critical unmet rate).

Two practical points follow. For measurement, evaluations of multilingual trustworthiness should report the fulfillment and correctness of critical content in addition to its presence. For model comparison, the proprietary model showed no language gap, while both open-weight models showed gaps of similar size. The study cannot attribute this difference to scale, training data or deployment, because the panel has only three models and their inference settings differ. In particular, Sol answered with medium reasoning effort while Qwen and Gemma ran with thinking disabled, and reasoning may be part of what lets a model complete a Bengali answer. A post hoc run with thinking enabled for Qwen and Gemma, on 30 tasks weighted toward hard ones, does not settle this. On those tasks the BN \ensuremath{-} EN critical-coverage gap moved from \ensuremath{-}0.127 to +0.021 for Qwen, partly because English coverage fell, and from \ensuremath{-}0.082 to \ensuremath{-}0.051 for Gemma; neither change is distinguishable from zero (Qwen +0.148, 95\% CI \ensuremath{-}0.008 to +0.344), and the thinking runs also used sampled rather than greedy decoding. Weaker Bengali performance in 9--12B open-weight models is not surprising in itself; what the study adds is where that weakness sits.

\textbf{Tokenization premium.} Bengali requires more tokens for the same task, consistent with the tokenization audit of Roy et al. \cite{roy2026tea}. In our runs, the Bengali prompt used about 1.75 times the input tokens of the English prompt for Sol, 2.00 for Qwen and 1.24 for Gemma (median of per-task ratios). These figures include a fixed system instruction and so understate the prompt premium. Bengali responses used 1.79, 1.75 and 0.98 times the visible output tokens of English, with Sol's reasoning tokens excluded so that the three models are compared on the same basis, and 1.72, 2.03 and 1.18 times the tokens per word. A Bengali user of Sol or Qwen therefore pays a real premium in cost and context. Across these three models, however, the premium does not predict the quality gap: Sol had a large premium and no coverage gap, while Gemma had the smallest premium and a clear gap. Nor did the premium cause truncation, since only 2 of 1,413 responses reached the output limit. With three models this comparison is descriptive, but it suggests that token cost and response quality are separate channels of disadvantage that should be measured separately.

The judge-validation record is also a result in its own right. A procedure with high RIU agreement (alpha 0.96) still failed safety-check agreement, and each disputed case traced to a gap in the instructions rather than to an error in the references. The study responses showed a similar pattern: agreement on RIUs held, while the unsupported-content check, which depends on a materiality standard, did not reach exact agreement. Requiring exact safety-check agreement, and building the held-out set around counterexamples to the revised rules, kept scoring from starting with an instrument that would have misclassified safety-relevant errors.

\section{Limitations}
\label{sec:limitations}

\begin{itemize}
\item \textbf{Judge.} All scores come from one automated judge, validated on constructed fixtures and then checked on 60 study responses. In that sample, agreement met the RIU thresholds but not the exact-agreement threshold for the unsupported-content check (0.933), so composite success outcomes that include the check inherit that error. Fifteen of the 60 responses were selected for low critical scores, which makes the pooled agreement and the error classification specific to this sample. The judge shares a provider with one generator. Its offset from the reviewer rating was the same in all three languages and the reviewer ratings agreed on every sampled Sol item, but Bengali RIU agreement (alpha 0.77) fell below the threshold, although on the stratified sample the reviewer's Bengali gap was larger than the judge's. The reason codes were checked only post hoc, by one reviewer, on 81 units.
\item \textbf{Held-out deviations.} The held-out set contains 88 of the 90 planned fixtures, and six references were revised after the judge pass through a blind re-rate limited to disputed fixtures.
\item \textbf{Multiplicity and equivalence.} Secondary contrasts and strata are unadjusted for multiplicity. No equivalence margin was set for the primary estimand, so the null result cannot be read as equivalence, and it is sensitive to the domain mix (Section~\ref{sec:results}).
\item \textbf{Sample.} The 157 tasks come from the pilot and calibration-expansion rounds and are not a random draw from the corpus, and each gold rubric had one independent reviewer. Validation rests on three reviewers, and every stage except calibration had a single rater, so reviewer-specific effects cannot be separated from the rating standard. The author drafted all material the reviewers checked. The English prompt is the canonical specification, so the Bengali conditions are carefully matched renderings and not queries composed by Bengali users. Prompt equivalence and criticality labels remain possible sources of measurement error.
\item \textbf{Panel.} The panel is small, the self-hosted and API-served models ran under different conditions, including the reasoning settings noted in Section~\ref{sec:discussion}, and each cell has one generation, so run-to-run variance is unknown and the effects should not be generalized to other models, languages or tasks.
\end{itemize}

\section{Conclusion}
\label{sec:conclusion}

Across 1,413 responses from three models on 157 matched EN/BN/BN-CS tasks, we detected no difference between Bengali and English in the outright omission of critical information, and omission was near zero throughout. Fulfillment did differ. For the two open-weight models, Bengali responses met fewer critical requirements in full, completed fewer tasks under the strict criterion and were more often scored incorrect or insufficient on critical content, while code-switching recovered about half of the loss in RIU coverage. Evaluations of trustworthiness across languages should therefore measure whether critical content is correct and sufficient, in addition to whether it is present.

\section{Availability}
\label{sec:availability}

Prompts, rubrics, validation fixtures and references, all 1,413 study responses with judge ratings, the reviewer QA ratings, analysis outputs, the robustness checks and the numbered pipeline scripts are released at
\begin{center}
\small\url{https://github.com/HeyAvijitRoy/multilingual-trustworthiness-study}
\end{center}
\noindent Given the score files, the analysis is deterministic (fixed bootstrap seed 20261023). Re-running generation requires a paid API for Sol and a CUDA GPU for Qwen and Gemma, and re-running scoring requires a paid API. Because responses and scores are included, no re-run is needed to reproduce any table in this paper. The frozen analysis plan and its table of deviations are in \texttt{PROTOCOL.md}.

\section{Risks and societal impact}
\label{sec:risks}

The study measures a gap that could harm Bengali speakers, who may act on answers that are present but incomplete or wrong. Our findings support more careful evaluation and are not a basis for deployment decisions, since three models, one language pair and a non-random task sample cannot support claims about any model's safety for Bengali users. Publishing the rubrics and responses carries little misuse risk because the tasks are ordinary help requests, although the scores should not be used to rank vendors. A single automated judge shaped every score, and the QA sample shows that it is not free of error, so absolute levels should be read with that error in mind. We used no human-subject data beyond the reviewers' ratings of model text.

\begin{acks}
We thank Proma Roy and two further native Bengali-speaking reviewers,
who preferred not to be named, for validating the prompts, rubrics and
scores reported in this paper. Proma Roy is a co-author of earlier work cited here \cite{roy2026,roy2026tea} and did not take part in preparing the
materials she reviewed.
This work used Jetstream2 at Indiana University \cite{jetstream2} through
allocation CIS260616 from the Advanced Cyberinfrastructure Coordination
Ecosystem: Services \& Support (ACCESS) program \cite{access2023}, which is
supported by U.S. National Science Foundation grants \#2138259, \#2138286,
\#2138307, \#2137603 and \#2138296.
\end{acks}

\section*{GenAI Usage Disclosure}

\begin{itemize}
\item \textbf{Studied models:} GPT-6.1 Sol, Qwen3.5-9B and Gemma 4 12B generated the analyzed responses; Gemini 3.8 Flash generated 20 unanalyzed exploratory responses.
\item \textbf{Automated judge:} GPT-5.4 (\texttt{gpt-5.4-2026-03-05}) scored responses and validation fixtures under the procedure described in Section~\ref{sec:judge}.
\item \textbf{Coding and writing assistants:} Claude Code assisted in drafting prompt realizations, rubrics, and validation fixtures, as well as supporting code verification and documentation. Codex was used to review the 17 QA disagreements discussed in Section~\ref{sec:qa}, while Grammarly assisted with grammar and readability. The author reviewed all AI-generated content. Independent reviewers, as described in Section~\ref{sec:setup}, validated the corpus, rubrics, and scores without participating in the preparation of the materials they evaluated.
\end{itemize}

\bibliographystyle{ACM-Reference-Format}

\bibliography{references}

\end{document}